# When the Knowledge Base Becomes the Gold Standard: Measuring Resource-Shared Evaluation Loops in Entity-Level Machine Translation

JINHYUNG BAE · DAIN KIL · SEONGMIN OH · SEUNGMIN LEE

*Hankuk University of Foreign Studies*

---

## Abstract

The *Seungjeongwon Ilgi*, a UNESCO Memory of the World record, is only 37.4% translated, and the most conspicuous failure mode in automatic translation is the person name — a misread name corrupts the historical fact rather than merely the surface. Low-resource historical domains have no expert gold standard for entity translation, so practitioners **substitute a knowledge base (KB) for the gold**. That KB is the same resource injected into the system: scoring becomes self-referential and the metric measures instruction compliance rather than translation quality.

We **measure** this loop. Using expert person-name annotations from the National Institute of Korean History as a gold independent of the injection pipeline, we hold the entity set fixed and vary only the provenance of the correct reading. Of 527 expert-annotated mentions, only **31.1%** lie outside the injection pipeline, and the residual loop is not uniform — in the overlapping segment the injected reading agrees with the human translation 97.8% of the time against 70.1% in the independent one, so **the segment that looks healthiest is the one the loop is holding up**.

Across four models, a difference-in-differences analysis shows the gain from KB injection is **confined to the segment whose gold shares the injected resource**; in the independent segment it is at or below zero. Post-injection preservation clusters in a narrow 0.910–0.996 band even though baseline capability differs fivefold, so the reported gain is the complement of prior performance and **weaker models appear to improve more dramatically**. On an independent sample built by removing the construction filter, the measure replicates within model (overlapping intervals) while discriminating between models (non-overlapping intervals) — it reflects a property of the model, not of the sample.

---

## 1. Introduction

### 1.1 One worked example

The `hanja` library renders 沈 as *chim* and 金 as *geum* — the readings for “to sink” and “metal.” As surnames the correct readings are **Sim** and **Kim**, and 金 is the most common Korean surname.

```
Source      ○ 右相沈悅四度呈辭。答曰, 安心調理。
Injected    [Persons — use exactly these Korean names]    沈悅 → 침열
(Chim-yeol)    ← WRONG
Human       우상 심열이 …              Sim-yeol             ← correct
Scored      "침열" ∈ output  →  counted correct
```

The system is **ordered** to emit a wrong name, and a gold standard built from **the same library** certifies it. Only comparison against the human translation exposes the error. This single case states the paper's problem.

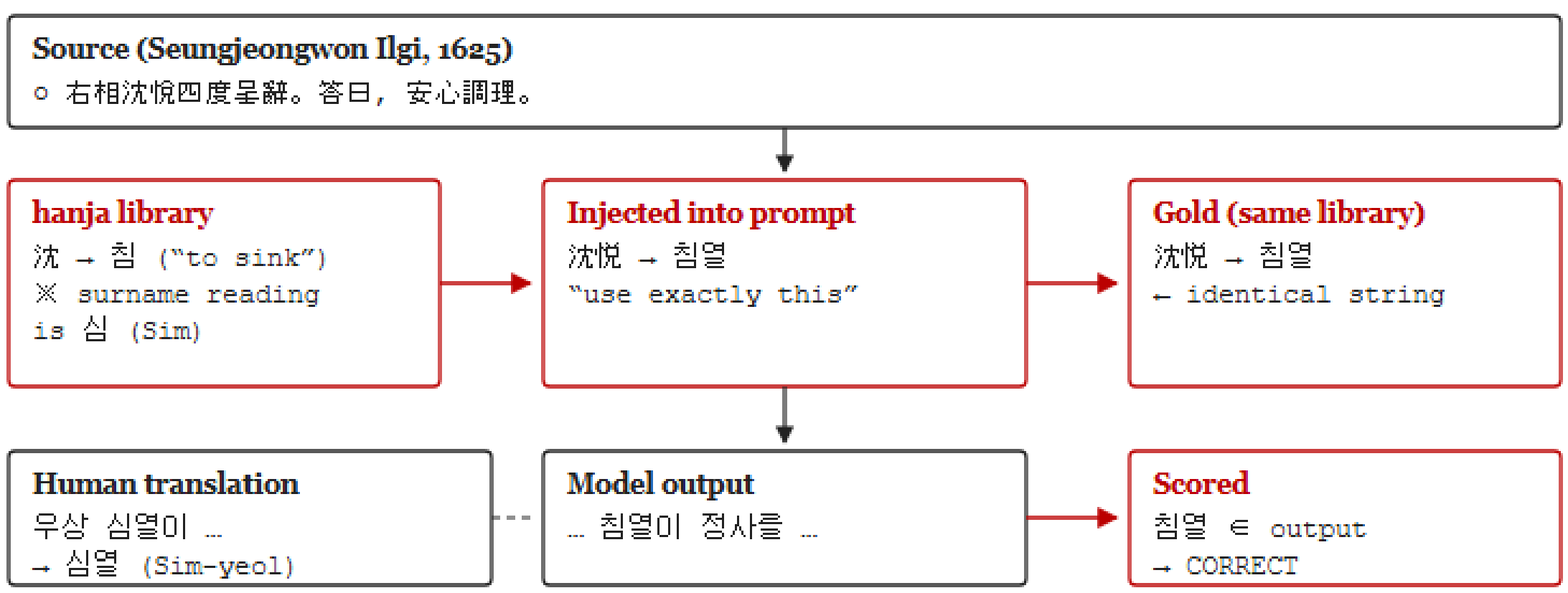

**Figure 1.** A wrong reading is injected, then certified by a gold built from the same library. The loop never leaves the library, so the error is invisible from inside it.

### 1.2 Background: a record whose translation is projected to finish in 2062

The *Seungjeongwon Ilgi* is the daily administrative journal of the Royal Secretariat of Joseon, kept without interruption from 1623 to 1910. It comprises 3,243 volumes and roughly 242.5 million characters, and as a primary source spanning the politics, diplomacy, economy and society of the later Joseon period it is inscribed in the UNESCO Memory of the World register.

The difficulty is access. Only 37.4% has been rendered into modern Korean, and because the supply of readers trained in classical Sinitic is limited, **completion at current throughput is projected for 2062**. The Institute for the Translation of Korean Classics has operated a neural machine translation system since 2017 with the explicit aim of bringing that date forward.

The bottleneck, however, is not sentence structure but **proper nouns**. A large share of entries follow formulaic administrative patterns (啓曰, 傳曰, 除授), so the syntax repeats; the person names and official titles inside them do not, and they are almost absent from general pretraining corpora. As §1.1 showed, Han-character names carry a polyphone problem: transliterated without context, they yield names that do not exist.

**The diagnosis that evaluation is unresolved comes from within the Korean field itself.** Comparing Baidu's translator with that of the Institute for the Translation of Korean Classics, Kim (2021) lists, among the tasks required to improve classical-Chinese machine translation, the construction of a cloud translation platform, the expansion of high-quality parallel corpora, and — explicitly — **the development of reliable translation evaluation methods**. This paper targets that gap, not by adding a metric but by measuring what the metrics in current use actually record.

It is therefore natural to inject a person-name knowledge base into the translation system. Doing so through the prompt, without fine-tuning, is especially attractive because an institution with no GPU infrastructure can adopt it immediately. This paper's concern is **not whether that method works**, but how its effect ought to be measured.

### 1.3 The problem: with no gold, what does one score against?

Scoring entity translation requires an answer key — which names occur in this source, and what is the correct rendering of each. In high-resource settings, human annotators build it. In a low-resource historical domain there is no budget for that.

The practical solution is to **build the key from the resources one has**: extract names with an NER model, attach Korean readings from a person-name knowledge base. This is reasonable, widely done, and there is no alternative.

But those resources are **the same ones injected into the translation system**. Strings drawn from the KB go into the prompt, and strings drawn from the same KB are used to score. What the metric then measures is not translational accuracy but **how faithfully an instruction was followed**. Because an injected name counts as correct simply by surviving into the output, even a **wrong** name is certified as correct — as *Chim-yeol* was in §1.1.

### 1.4 Contributions

We propose no new metric, and we do not claim to have discovered circular evaluation bias. Reference bias (Fomicheva & Specia, 2016), pseudo-reference bias (Albrecht & Hwa, 2007), LLM self-preference (Panickssery et al., 2024), benchmark contamination, and evaluation circularity in retrieval-augmented generation (Dietz et al., 2025; 2026) all precede us. Entity accuracy metrics likewise exist in KoBE (Gekhman et al., 2020) and M-ETA (Conia et al., 2025).

What we do is **measure**.

1. **We quantify how independent the gold actually is.** Of 527 expert annotations, 31.1% lie outside the injection pipeline; for NER Recall the figure is 0%. To our knowledge this quantity has never been reported for entity-level MT evaluation.
2. **We show what the inflation is a function of.** Its magnitude is set by KB coverage and model capability; in particular, the post-injection ceiling is near-invariant across models, which makes the reported gain the complement of prior performance.
3. A ***Seungjeongwon Ilgi* case study**: four models, three conditions, a stratified sample of 300 documents, with all model outputs released.

Hardening the scoring protocol (over-generation penalty, positional check) is left to §10 as future work. All measurements here use the **same presence-based scoring that prior work uses** — changing the metric would confound whether the observed inflation comes from the protocol or from the loop.

---

## 2. Related Work

### 2.1 NLP for Han-character historical records

Computational work on Joseon-period Sinitic records accumulated rapidly through the 2020s. **HUE** (Yoo et al., Findings of NAACL 2022) released a benchmark bundling chronological attribution, topic classification, named entity recognition and summary retrieval over the Annals of the Joseon Dynasty and the *Seungjeongwon Ilgi*, and showed that BERT-family models continued-pretrained on those two corpora substantially outperform general models. **H2KE** (Son et al., 2022) approached Hanja→Korean and Hanja→English translation with multilingual seq2seq models, and **HERITAGE** (Song et al., 2025) integrated punctuation restoration, entity recognition and translation into a single web platform.

> **The delta with respect to HUE should be stated plainly.** HUE treats the *Seungjeongwon Ilgi*, Hanja and entity recognition **as tasks**. What we study is not entity-recognition performance but **what happens when that output is injected into a translation prompt and then used to score it.** The corpora overlap; the question does not.

There is a Korean lineage as well. Yu et al. (2021) partition Korean NER errors into detection, boundary, segmentation and labelling types, showing per-label error rates ranging from 1% to 30% — an error typology that connects directly to the protocol hardening discussed in §10.7. Bae et al. (2019) analyse error patterns in classical-Chinese machine translation by punctuation availability and difficulty, and set out the prerequisites for AI translation research in the field.

In translation-evaluation methodology, Jeong (2018) frames the adequacy of translation assessment in terms of **validity, reliability and practicality**, and examines whether automatic metrics can be applied to human translation. The reliability / discriminant-validity framing we use in §5.6 shares its vocabulary with that lineage. Korean work on LLM translation quality since 2023, however, remains largely native-speaker qualitative assessment or sentence-level error classification; **we found no case treating entities under a quantitative protocol.**

On cross-lingual transfer, **Song et al. (2025, "Shared Heritage, Distinct Writing")** report that transferring Classical Chinese resources to Han-character processing yields statistically negligible gains (±0.0068 F1 for NER). Our observation in §6.1 is stronger: an 8B model carrying a Classical Chinese corpus scored **below** a 2B model without one on entity accuracy.

### 2.2 Entity-aware machine translation and its evaluation

Treating entities as a first-class problem in MT is by now mature. **SemEval-2025 Task 2** (Conia et al.) organised entity-aware machine translation as a shared task across ten language pairs and 34 systems, adopting **M-ETA** (Manual Entity Translation Accuracy), grounded in human gold translations, as the official measure. Knowledge-base injection, retrieval augmentation and self-refinement — the family our method belongs to — were all standard entries there.

On the evaluation side, **KoBE** (Gekhman et al., Findings of EMNLP 2020) proposed scoring without reference translations by grounding entities from both source and candidate in a large multilingual knowledge base and measuring recall. Crucially, KoBE already diagnosed that **a recall-only measure inflates when a system over-produces entities** and introduced an Entity Count Penalty. **Alam et al. (2021)** made the point adversarially for terminology consistency: a "cheating" system that simply appends the required term to the end of its output is not caught at all by presence-based scoring.

> This paper **proposes no new metric.** Our scoring is the same presence-based rule prior work uses, and its weaknesses were identified by the two studies above. Changing the metric would confound whether the observed inflation came from the protocol or from the loop, so we deliberately keep the scoring fixed (§10.7).

### 2.3 Circularity in evaluation

The recognition that scores distort when evaluation resources overlap system resources is long established in MT. **Fomicheva & Specia (ACL 2016)** documented reference bias: shown a reference, monolingual annotators are pulled toward it. **Albrecht & Hwa (ACL 2007; WMT 2008)** showed that using other systems' outputs as pseudo-references favours systems of the same family. More recently **Panickssery et al. (NeurIPS 2024)** quantified self-preference: LLM judges recognise and favour their own generations. The benchmark-contamination literature covers the parallel case in which evaluation answers overlap training data. Most recently the structure has been named directly in retrieval-augmented generation: **Dietz et al. (ECIR 2026)** show that when components of the evaluation method are integrated into the RAG system itself, scores inflate and agreement with manual assessment drops; the same group lists "circularity — leaking the evaluation signal into the system" first among their LLM-evaluation tropes (Dietz et al., 2025). Those studies concern shared nuggets and judge prompts in long-form generation; we specialise the structure to entity-level MT, where the shared resource is a person-name KB, and quantify how much of the gold it reaches (§4).

The **CALBC Silver Standard Corpus** in biomedical text mining shows a classical remedy: the key was harmonised from participating systems' outputs, but **a participant that had trained on one round's key was excluded from generating the next**. Resource sharing was blocked procedurally.

> Our setting is the **extreme case** of this lineage. Reference bias concerns a key that carries one translator's idiom; self-preference concerns a judge recognising its own text. Here the key is **generated from the very resource injected into the system.** And this arises not from carelessness but as a structural consequence of working where expert gold cannot be purchased (§9.1).

### 2.4 A note on terminology

The phrase "circular evaluation" is already taken. The multimodal benchmark **MMBench** (Liu et al., ECCV 2024) uses **CircularEval** for a protocol that rotates multiple-choice options and

rescores — a robustness procedure unrelated to resource sharing. To avoid confusion we use **resource-shared evaluation loop** throughout.

### 2.5 Where gold exists, the problem does not arise

A contrast case sharpens the condition. The ancient-Chinese NER evaluations **GuNER 2023** (CCL) and **EvaHan 2025** (ALP) both presuppose manual annotation by people trained in classical Chinese, and both note explicitly that this annotation is expensive. Because the gold can be bought, there is no reason to substitute system resources for it, and the conditions for a loop never form. **Circularity is not methodological negligence; it is a function of budget.**

---

## 3. Experimental Design

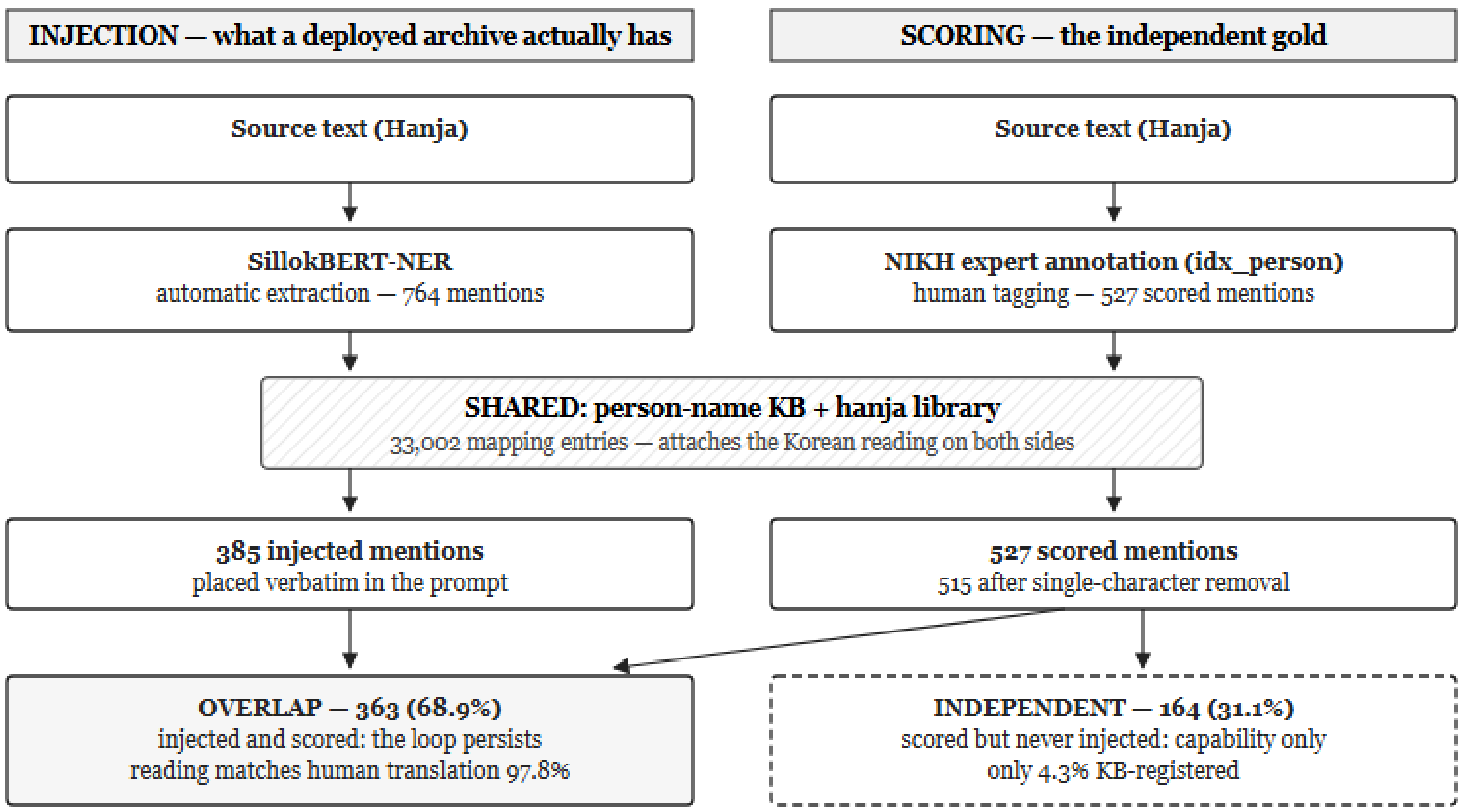

**Figure 2.** The injection and scoring pipelines, and the segment where they still share a resource. The entity sets are separate but the resource supplying the Korean reading is not, which is why the loop survives on 68.9% of the gold.

### 3.1 Separating the two pipelines

The system and the evaluation must not share resources. Our configuration:

| | Entity set | Korean reading | Role |
|---|---|---|---|
| **Injection** | SillokBERT-NER over the source | KB (`person_master` + inverted index) | what a deployed archive actually has |

| | Entity set | Korean reading | Role |
|---|---|---|---|
| **Scoring** | NIKH expert `idx_person` tags | KB, else `hanja` transliteration | independent gold |

This separation is an operational fact before it is a methodological requirement: **an institution that already held expert person-name annotations would not need a translation aid at all.** What a deployable pipeline has is an NER model and a KB.

> In the original experiment two scripts wrote to the same filename, so the injection gold was overwritten by the scoring gold. The separation therefore held **by accident**, and the injection gold was lost. We reconstruct it under a distinct filename, making the separation **deliberate**.

### 3.2 Evaluation set

From 62,476 parallel pairs of the Injo reign we drew 300 documents by proportional stratified sampling over length buckets (XS/S/M/L) × style patterns (GEN/REP/DEN/MEM/ROY/APT/APR), seed 42.

**A sentence-level BLEU ≥ 20 filter using the primary model was applied during construction.** Its purpose was not model selection but **removal of merge errors**. The corpus was built by joining the Annals and the *Seungjeongwon Ilgi* on date identifiers, so misaligned pairs occur in which the source and the human translation refer to different entries. Such pairs score low under any model, so a BLEU floor removes them.

**The filter cannot, however, distinguish misalignment from model difficulty**, since a low BLEU may indicate either. We therefore decompose its effect using an independent 300-document sample that retains the stratification and removes only the filter (§5.6).

### 3.3 Conditions

| Condition | Content |
|---|---|
| baseline | minimal prompt, source only |
| few-shot | five fixed style exemplars |
| NERfix | rule-based post-hoc name correction over few-shot output |
| **NERinject** | Hanja→Hangul name block inserted immediately before the source in the few-shot prompt |

The intersection between person names occurring in the few-shot exemplars and the evaluation gold is a single item (an incidental substring match). few-shot therefore conveys no substantive name information and is a **valid control** for the injection effect.

---

## 4. How Independent Is the Gold?

### 4.1 Injection set versus scoring set

| | Count | Share of scoring set |
|---|---|---|
| Scoring set (expert tags) | 527 | 100% |
| ├ **Overlap** — injected and scored | **363** | **68.9%** |
| └ **Independent** — never injected | **164** | **31.1%** |
| Injection-only (NER found, experts did not tag) | 22 | — |

**Two denominators are used in this paper and they must not be conflated.** All counts in §4 are stated over the **527** expert tags whose reading could be resolved. Entity accuracy (ETS), however, is scored over **515**: twelve single-character tags are excluded because a one-character string matches as a substring of unrelated words and would inflate the score. Under that exclusion the decomposition becomes overlap 360 and independent 155 — the nine of the twelve that fall in the independent segment are the reason the overlap share moves from 68.9% to 69.9%. The distinction changes no conclusion in this paper, but §5 tables are on 515 (or 361 when restricted to the four-model common set of 240 documents) while §4 tables are on 527.

**NER Recall scores against the injection set itself: independence 0%.** Replacing it with expert annotations recovers 31.1%.

### 4.2 The loop closes where the KB is strong

| Segment | n | KB-registered | Reading matches human translation |
|---|---|---|---|
| Overlap (loop persists) | 363 | 100% *(by construction)* | **97.8%** |
| Independent | 164 | **4.3%** | **70.1%** |

The first column is not an empirical finding on the overlap row: injection requires a KB reading, so every overlapping mention is KB-registered **by construction**. The finding is on the other row — only **4.3%** of the independent segment is KB-registered. What the NER model misses is precisely what the KB lacks, so the two segments are structurally opposite: mentions an NER model finds *and* experts tag are well-attested figures, while those only experts tag are obscure ones absent from the KB.

The second column carries the weight, because it is not definitional. The injected reading agrees with the human translation for **97.8%** of the overlapping segment but only **70.1%** of the independent one — a gap of 27.7 points in the fidelity of the answer key itself.

> **The segment that looks healthiest is the one the loop is holding up.**

### 4.3 Bounding the distortion

Disagreement between the readings used for injection/scoring and those actually used by the human translation:

| Segment | n | Disagreement |
|---|---|---|
| KB-registered | 370 | **2.2%** |
| `hanja` transliteration fallback | 157 | **31.2%** |
| **Total** | **527** | **10.8%** |

All of the distortion sits in the KB-absent segment. Linear projection gives 10.8% at 70% coverage, 16.7% at 50%, 22.5% at 30% and 28.3% at 10%. **An archive can estimate its own exposure from its KB coverage alone.**

---

## 5. How Much Does the Loop Inflate?

### 5.1 A three-level decline under changing gold provenance

The same outputs, scored against golds of differing provenance (Qwen3-8B, NERinject):

| Gold provenance | Loop | Value |
|---|---|---|
| NER Recall (injection set = scoring set) | complete | **0.944** |
| ETS overall (expert tags) | partial | **0.765** |
| ETS independent (expert tags − injection set) | none | **0.348** |

The baseline independent segment scores 0.342, so **under an independent gold NERinject is effectively indistinguishable from baseline.** The reportable figure is 2.7× the honest one.

### 5.2 Difference-in-differences

To test whether the gain is confined to the loop segment we difference the two segments:

```
loop contribution = [NERinject(overlap) − few-shot(overlap)]
                  − [NERinject(independent) − few-shot(independent)]
```

few-shot, not baseline, is the reference: the NERinject prompt contains the few-shot exemplars verbatim, so a baseline contrast would **confound the style effect**.

On the 240 documents common to all models and conditions (overlap n=244, independent n=117):

| Model | BLEU(c) | baseline ETS | Δ overlap | Δ independent | loop contribution |
|---|---|---|---|---|---|
| **Qwen3-4B** | 5.21 | 0.133 | +0.697 (+170/244) | −0.026 (−3/117) | **+0.722** |
| **Qwen3-8B** | 15.57 | 0.374 | +0.656 (+160/244) | −0.026 (−3/117) | **+0.681** |
| **gemma-3n-E2B** | 14.99 | 0.515 | +0.389 (+95/244) | −0.026 (−3/117) | **+0.415** |
| **gemma-4-26b** | 34.57 | 0.723 | +0.225 (+55/244) | −0.068 (−8/117) | **+0.294** |

> The identical −0.026 across three models is **coincidental**: each lost a net three of the 117 independent entities (−3/117 = −0.02564). Counts are given alongside so the arithmetic is visible. BLEU(c) and baseline ETS here are computed on the 240 common documents and therefore differ from the full-300 figures in §6.1 (BLEU 14.07 / 13.97 / 34.41; ETS 0.396 / 0.499 / 0.723). The monotone relation in the loop contribution holds under either denominator.

**Correlation between baseline ETS and loop contribution: r = −0.938.**

**The statistical backing comes from McNemar's exact test.** Comparing few-shot with NERinject entity-by-entity, the overlap segment is overwhelmingly significant for all four models — improvement-to-regression counts of 171:1, 160:0, 98:3 and 56:1, with p from 5.8e-50 to 8.1e-16. In the independent segment **three models are non-significant** (3:6, p=0.508) and **gemma-4-26b is significantly worse** (0:8, p=0.008). No model shows an improvement there. The confinement of the gain is thus established by paired testing rather than by a correlation coefficient: with n=4 we report r for direction and magnitude only, while the abstract's claim rests on these tests.

The 95% confidence intervals for the difference-in-differences (document-level cluster bootstrap, B=1000) are [+0.619, +0.812], [+0.598, +0.774], [+0.335, +0.495] and [+0.189, +0.437] respectively. **The upper bound for gemma-3n and the lower bound for Qwen3-8B do not overlap**, so the difference between models is not explained by sampling variation.

Qwen3-4B also serves as a **consistency check** on the account in §5.3. If the ceiling is near-invariant, then the loop contribution is fixed by prior performance alone. On this model's own completed set (n=300) the pre-injection overlap was 0.225, so a ceiling of 0.95 implies +0.725; the measured value was **+0.681**, the realised ceiling being 0.906 rather than 0.95. The account therefore recovers the magnitude to within 0.044, with the residual dominated by the ceiling assumption.

We make **no claim of priority** for this calculation. It was written up in the same commit as the result it describes, so the repository cannot establish that it preceded the measurement, and we do not present it as a pre-registered prediction.

Computed on each model's own completed set the monotone relation is preserved. **The conclusion is invariant to the denominator.**

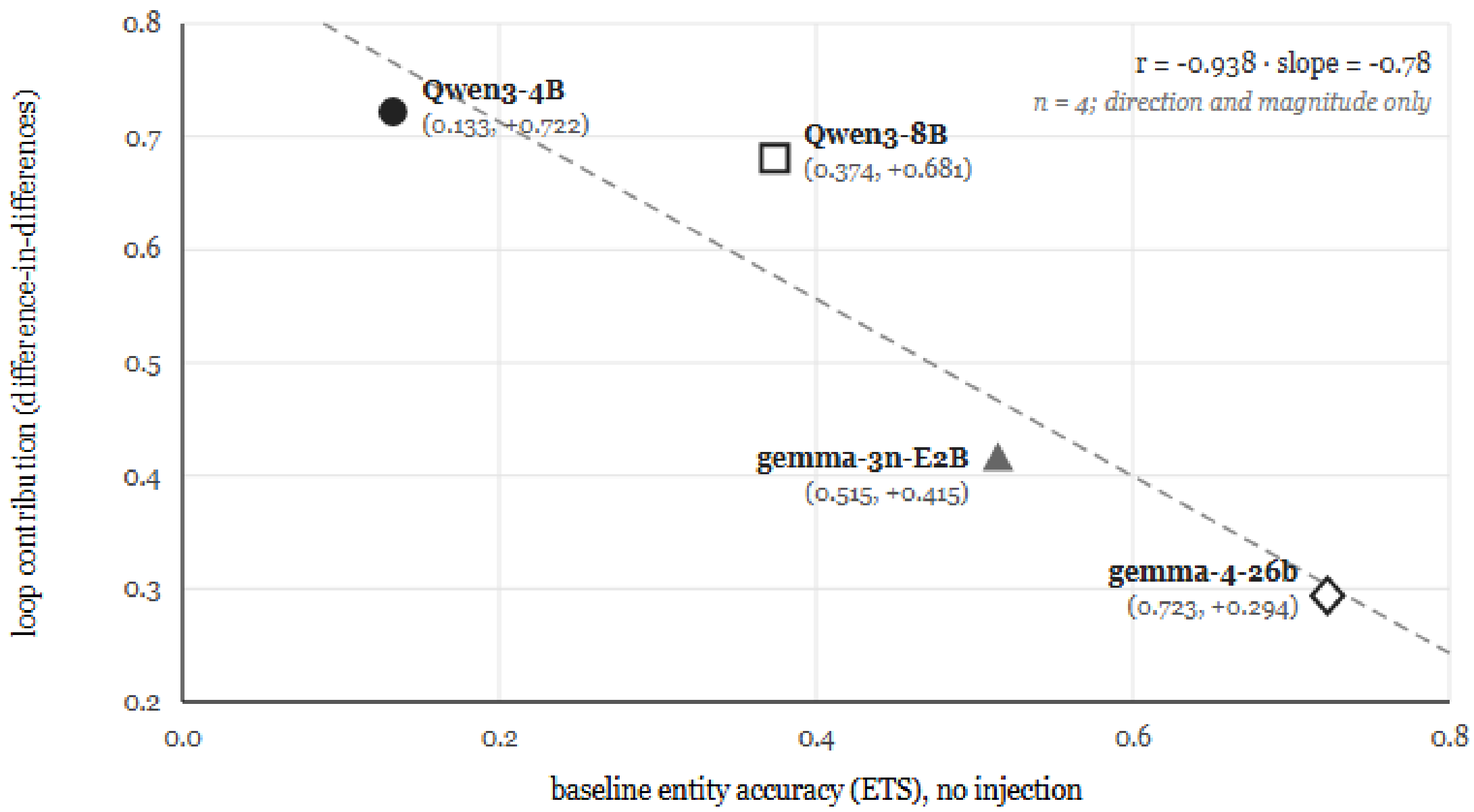

**Figure 3.** The weaker the model, the larger the loop's contribution (four models, common 240-document set). With n=4 we report direction and magnitude only; monotonicity survives dropping the extreme point (Qwen3-4B).

### 5.3 Ceilings cluster in a narrow band

| Model | baseline ETS | overlap, pre-injection | **overlap, post-injection (ceiling)** |
|---|---|---|---|
| Qwen3-4B | 0.133 | 0.213 | **0.910** |
| Qwen3-8B | 0.374 | 0.303 | **0.959** |
| gemma-3n-E2B | 0.515 | 0.590 | **0.980** |
| gemma-4-26b | 0.723 | 0.770 | **0.996** |

The ceiling rises modestly with capability but occupies a **very narrow band**. Even the weakest model reaches 0.910, and it goes no higher — apparently limited by instruction-following ability itself.

| | Range | Width |
|---|---|---|
| Ceiling | 0.910 – 0.996 | **0.086** |
| Pre-injection | 0.213 – 0.770 | **0.557** |

**Pre-injection performance varies 6.5× more than the ceiling.** The reported gain is therefore governed by prior performance rather than by the ceiling, and weaker models appear to improve more dramatically. This measures instruction compliance, not translation.

*(Note: `Δoverlap = ceiling − pre` is an identity by definition. What is tested is that the ceilings **cluster in a narrow band**; the complement structure follows.)*

### 5.4 Spillover is not zero but negative

The independent segment is **at or below zero for all four models**: −0.026 (−3/117) for Qwen3-4B, Qwen3-8B and gemma-3n-E2B, and −0.068 (−8/117) for gemma-4-26b.

Injection does not generalise beyond the injected list and, in the stronger model, is **mildly harmful**: attention to the injected names appears to come at the expense of the remaining ones. That the spillover is negative rather than zero was not predicted, and it answers the objection that "of course there is no effect where nothing was injected" — **were the result trivial, the sign would not be negative.**

### 5.5 Swapping the reading provenance does not change the gain — a fixed-denominator contrast

Circularity has two axes: the **set** axis (what is scored) and the **reading** axis (what counts as correct). §5.2 measured the set axis. Here we hold the denominator fixed and vary **only the provenance of the reading**.

- **ETS-KB**: scored against the same strings the injection uses (KB / `hanja`) → partial loop
- **ETS-REF**: scored against the wording actually used by the human translation → no loop

The denominator is fixed at the 355 entities for which both golds exist. The reference gold was settled at 506/515 (98.3%): 494 by automatic matching, 10 by rule extension (initial-sound rules, surname exceptions, title-suffix stripping, spacing variants), 2 by individual adjudication. Five cases where the human translation does not name the person (REF-absent) and four where the tag and the translation refer to different people were excluded.

| Model | Δ ETS-KB | Δ ETS-REF | **Difference (inflation)** |
|---|---|---|---|
| Qwen3-4B | +0.468 | +0.473 | **−0.006** |
| Qwen3-8B | +0.442 | +0.459 | **−0.017** |
| gemma-3n-E2B | +0.256 | +0.259 | **−0.003** |
| gemma-4-26b | +0.130 | +0.135 | **−0.006** |

**Reading provenance does not distort the gain.** All four models fall within 0.02 and the sign is inconsistent. The inflation arises entirely on the set axis.

It does, however, **distort the absolute level.** Under NERinject, ETS-KB sits below ETS-REF:

| Model | ETS-KB − ETS-REF |
|---|---|
| Qwen3-4B | −0.017 |

| Model | ETS-KB − ETS-REF |
|---|---|
| Qwen3-8B | −0.020 |
| gemma-3n-E2B | −0.048 |
| **gemma-4-26b** | **−0.068** |

**The gap widens with capability.** A capable model produces the correct reading (*Sim-yeol*) while the KB gold expects the wrong one (*Chim-yeol*) and marks it incorrect. KB-derived golds therefore penalise the more competent model more heavily.

> **Practical guidance**: a KB-derived gold is usable for within-condition deltas, but it **understates absolute performance** in proportion to model capability and is therefore **unsuitable for cross-model absolute comparison.**

### 5.6 Decomposing the evaluation-set filter: merge errors or difficulty selection?

To test whether the BLEU ≥ 20 filter drove our results, we constructed an independent 300-document sample that **retains the stratification and removes only the filter** (population 62,476 pairs; reference-deduplicated; original evaluation ids excluded; proportional stratified sampling over length × style cells; seed 20250729). The id intersection between the two samples is zero.

**The two samples are structurally almost identical.**

| | Original set | Unfiltered control |
|---|---|---|
| Length XS/S/M/L | 133/104/43/20 | 135/99/45/21 |
| Scoring entities | 527 | 557 |
| Overlap (loop) | 68.9% | 68.2% |
| **Independent** | **31.1%** | **31.8%** |

That the independence rate is 31.1% against 31.8% suggests the figure in §4.1 is not an artefact of the particular sample.

**Misalignment rates show the filter did clean the data.** Treating a pair as a misalignment candidate when its translation/source length ratio is extreme relative to the median (below 0.35× or above 3×):

| | Misalignment candidates |
|---|---|
| Original set (filtered) | **0 / 300 = 0.0%** |
| Unfiltered control | **14 / 300 = 4.7%** |

A typical case has a source recording appointments in the Office of the Inspector-General while the translation describes a criminal case in the State Tribunal — **two entirely different entries**. This is a by-product of date-identifier joining and is unrelated to translation quality.

**The decomposition behaves differently by metric.** Comparing the baseline condition on the filtered and unfiltered samples (each on its own full 300):

| Model | BLEU(c) | ETS overall | ETS overlap | ETS independent |
|---|---|---|---|---|
| Qwen3-8B | **−2.79** | −0.045 | −0.051 | **−0.031** |
| gemma-3n-E2B | **−2.02** | **+0.034** | +0.069 | **−0.041** |

**(i) BLEU falls consistently for both models, and most of the fall is merge error.** Removing the 14 misaligned pairs raises Qwen3-8B from 11.29 to 13.59, so 2.30 of the 2.79 gap (83%) disappears and the residual is 0.48. The filter performed its intended cleaning function.

| Qwen3-8B baseline | BLEU(c) | ETS overall |
|---|---|---|
| Unfiltered, all (n=300) | 11.29 | 0.351 |
| Unfiltered, minus 14 misalignments (n=286) | **13.59** | 0.343 |
| Original set (n=300) | 14.07 | 0.396 |

**(ii) ETS moves in opposite directions for the two models** (−0.045 versus +0.034). There is therefore **no evidence that the filter systematically inflated entity accuracy.** This is consistent with the fact that ETS does not use the human translation: gold entities come from the NIKH HTML and scoring inspects only the model output, so source–translation misalignment cannot affect it. Indeed removing the misaligned pairs lowers Qwen3-8B's ETS slightly (0.351 → 0.343).

**(iii) Only the independent segment falls consistently for both models** (−0.031, −0.041). The unfiltered sample contains more obscure, KB-absent figures; this reflects the random sample being genuinely harder rather than a bias introduced by the filter.

> **Conclusion**: the filter cleaned the data, and most of the BLEU gap (83%) follows from that. No systematic bias in entity accuracy is observed. Absolute BLEU should be reported with the upward bias stated; claims about condition deltas are defended by the difference-in-differences replication below.

**The difference-in-differences replicates across samples while discriminating between models.**

| Model | Sample | Pre-injection overlap | Ceiling | Δ overlap | Δ independent | **DiD** | 95% CI |
|---|---|---|---|---|---|---|---|
| Qwen3-8B | original | 0.339 | 0.944 | +0.606 | +0.000 | **+0.606** | [+0.528, +0.682] |
| Qwen3-8B | unfiltered | 0.340 | 0.979 | +0.639 | +0.018 | **+0.621** | [+0.554, +0.685] |
| gemma-3n-E2B | original | 0.589 | 0.967 | +0.378 | −0.013 | **+0.391** | [+0.326, +0.463] |
| gemma-3n-E2B | unfiltered | 0.573 | 0.955 | +0.382 | −0.012 | **+0.394** | [+0.321, +0.464] |

**Denominator note.** This table uses each sample's **own full 300**. The four-model table in §5.2 uses the **240 documents common to all four models**, so values differ (e.g. Qwen3-8B on the original set: DiD +0.606 vs +0.681, ceiling 0.944 vs 0.959, Δ independent +0.000 vs −0.026). The monotone relation and the cross-sample replication hold under either denominator.

Two properties hold simultaneously.

- **Within-model reliability** — for a given model the two samples differ by 0.003–0.015 and the confidence intervals overlap. The components agree as well (gemma-3n's pre-injection overlap 0.589 vs 0.573, ceiling 0.967 vs 0.955, Δ independent −0.013 vs −0.012).
- **Between-model discrimination** — within a given sample the two models differ by 0.21–0.23 and the confidence intervals do not overlap. This holds in both samples.

The measurement reflects a property of the **model**, not of the sample. Satisfying both reliability (same object → same value) and discriminant validity (different objects → different values) rules out the explanation that the observed inflation is an artefact of sample selection.

This also answers the objection that "of course there is no effect where nothing was injected." Were the result trivial, the models would not be distinguishable; in fact their intervals separate in both samples.

**Style guidance was checked the same way.** From baseline to few-shot, BLEU rises consistently across all four model × sample combinations (+2.29 to +2.71), but for entity accuracy **only one of the four measurements is significant, and it does not replicate.**

| Model | Sample | ΔBLEU | Δ ETS | improve:regress | p |
|---|---|---|---|---|---|
| Qwen3-8B | original | +2.44 | −0.054 | 27:55 | **0.003** |
| Qwen3-8B | unfiltered | +2.71 | −0.017 | 38:47 | 0.386 |
| gemma-3n-E2B | original | +2.58 | +0.029 | 51:36 | 0.133 |
| gemma-3n-E2B | unfiltered | +2.29 | −0.018 | 42:52 | 0.353 |

The −0.054 (p=0.003) seen for Qwen3-8B on the original set becomes −0.017 (p=0.386) on the independent sample, and gemma-3n reverses sign between samples. **A trade-off claim resting on a single measurement is not supported under replication** (§6.2).

---

## 6. Secondary Findings

### 6.1 Classical Chinese pretraining does not transfer

| Model | Parameters | BLEU(c) | baseline ETS |
|---|---|---|---|
| Qwen3-8B | 8B | 14.07 | **0.396** |
| gemma-3n-E2B | 5B (2B effective) | 13.97 | **0.499** |

| Model | Parameters | BLEU(c) | baseline ETS |
|---|---|---|---|
| gemma-4-26b | 26B | 34.41 | 0.723 |

Qwen3-8B, which carries a Classical Chinese (文言文) corpus, scores **below** a 2B-class model without one. Sharing the Han script is not sufficient: the Korean readings of Joseon person names conflict with Chinese ones. This is a stronger statement than the “negligible transfer” observation of Song et al. (2025).

Because the capability axis does not track parameter count (2B > 8B), we regress against **measured baseline ETS** rather than model size.

### 6.2 Style guidance and entity accuracy are not a trade-off

For gemma-4-26b, few-shot raised BLEU by +5.40 while leaving ETS at **0.739** → **0.739**.[1] The trade-off reported in the original study (−0.012) does not reproduce.

**Extending to two models and two samples settles the question** (§5.6). Only one of the four measurements is significant (Qwen3-8B, original set, p=0.003), it fails to replicate on the independent sample (p=0.386), and the remaining three are non-significant with inconsistent sign. BLEU, by contrast, improves consistently in all four combinations (+2.29 to +2.71).

> **No trade-off between style guidance and entity accuracy is observed.** The original claim, based on a single measurement, is not supported under replication.

---

## 7. Limitations

- Prior-art coverage comprises ACL Anthology, arXiv, Springer, the public CCL/CNKI indices, and the KCI / DBpia / KISS web indices. Korean work on classical-Chinese MT, historical entity recognition and automatic translation evaluation was identified, but **no Korean precedent addressing circularity from shared evaluation resources was found.** As this rests on web indices, a residual risk of an unindexed precedent remains.
- **Absolute BLEU is not comparable to prior work.** Corpus (Annals/collections vs *Seungjeongwon Ilgi*), test set construction and BLEU tokenization all differ. We claim only within-set condition deltas.
- The evaluation set was built with a BLEU ≥ 20 filter. §5.6 decomposes its effect: merge-error removal accounts for 83% of the BLEU gap, and no systematic bias in entity accuracy is observed (the sign differs across models).
- The injection gold is a **reconstruction**; the original was overwritten by a same-named file. The structural separation reproduces, exact set identity does not.
- False positives are defined over Injo-period KB names. That dictionary holds **2,533** entries: the KB lists 2,690 persons for the Injo reign, of which we keep the distinct

[1] This 0.739 is computed on gemma-4-26b’s own baseline set (300 documents, 515 scored mentions). The 0.723 in §5.2 and §6.1 is the same model on the four-model common set (240 documents, 361 mentions). The denominators differ; neither value is in error.

Korean names of two or more syllables (single-syllable names cause substring false alarms). Insertions outside this dictionary escape detection.

- gemma-4-26b covers **240/300** across the three conditions. The conditions complete 300, 266 and 254 documents respectively, and the three-way intersection is 240; see §8.
- The unfiltered replication (§5.6) covers Qwen3-8B and gemma-3n-E2B. gemma-4-26b was excluded because the alias instability documented in §8 makes a re-run itself irreproducible, and Qwen3-4B because the same family is already represented by its 8B sibling. Since the difference-in- differences is a within-model paired contrast, two mid-sized models suffice to show that the value survives a structurally very different sample.
- The style comparison is blind but conducted by the authors, and is reported as preliminary.

---

## 8. Reproducibility as a Result, Not an Excuse

During this study the served behaviour of the alias `gemma-4-26b-a4b-it` changed. Under **identical code, parameters and prompts**, mean output length fell from 244 to 34 characters and 185 of 300 outputs became empty: the model emits `<thought>` reasoning whose tokens consume the output budget.

Reasoning cannot be disabled on this endpoint — `thinkingBudget`, `reasoning_effort` and `chat_template_kwargs` are all rejected with HTTP 400. The only lever is the token budget; raising it from 2,048 to 8,192 resolved most cases, but **34 documents did not complete even at 8,192.** Those documents consume over 19,000 characters on reasoning alone and **did not complete at 32,768 tokens within 500 seconds.**

The original run completed 300/300 at `max_tokens=2048`.

> **An unversioned commercial API alias does not preserve reproducibility even when the code is archived.** We report the alias, the access date and the token budget, and release all model outputs.

---

### Data Availability

Model outputs, evaluation code, and the document identifiers of the 300-document evaluation set (and of the unfiltered control sample of §5.6) are released at **https://github.com/nepersoned/malmoi-sjw-eval**. The Hanja source texts, Korean translations, and expert person-name annotations are copyright of the National Institute of Korean History (All Rights Reserved) and are not redistributed; we provide document identifiers and retrieval scripts so that these materials can be obtained directly from `sjw.history.go.kr`. An inquiry regarding formal permission for redistribution has been submitted to the institute.

Two consequences of that restriction bear on replication. The human translation was removed from every released output record. Where a model reproduced the source instead of translating it, runs of eight or more Han characters are masked, so **BLEU cannot be recomputed for those 189 records from the released files**; the BLEU figures reported here were computed before masking. Entity accuracy is unaffected, as it matches Korean readings. The unmasked outputs are available to reviewers on request.

---

## 9. Implications for Cultural-Heritage Practice

### 9.1 Why this arises only in low-resource archives

The chain runs: no fine-tuning means no GPU is needed; if the method works on free models, **an archive can adopt it at zero budget**. But a zero budget also means no money for an expert gold standard, so a knowledge base is substituted — and that is exactly where the loop appears.

This is not carelessness but a **structural consequence of resource constraints**. Where an expert gold does exist, the problem does not arise. The ancient-Chinese NER evaluations (GuNER 2023, EvaHan 2025) explicitly presuppose manual annotation by people trained in classical Chinese, and because that cost is paid, the conditions for a loop never form. Circularity appears only where the gold **cannot be bought** — which is precisely the setting we study.

### 9.2 How a misread name changes the historical record

An entity error is not a surface error. Two cases from our evaluation set show why.

沈悅**.** A civil official of the Injo reign. The source ○ 右相沈悅四度呈辭。答曰， 安心調理。 records that the Right State Councillor submitted a fourth resignation and the king declined it, urging him to rest. The `hanja` library renders the name *Chim-yeol*; the human translation reads *Sim-yeol*. **No person named *Chim-yeol* exists in the Joseon biographical record.** A researcher reading the machine translation cannot resolve the name, and cannot establish whose resignation this politically significant entry describes.

金瑬**.** The NIKH tag reads 昇平府院君金瑬, so transliterating the string with its title attached yields *Seungpyeong-buwongun Geum-ryu*; the human translation reads *Kim Ryu*. 金 is the most common Korean surname and reads *Kim*. Kim Ryu was a principal figure in the 1623 Injo Restoration; if his name is rendered *Geum-ryu*, this entry drops out of any search tracing the movements of the restoration's meritorious subjects.

What the two share is that **the misrendered names do not exist.** The error does not merely assert a falsehood; it removes the entry from the net of archival search. For a programme whose goal is complete translation, the result is text that has been translated but cannot be used.

### 9.3 A procedural checklist for archives

1. **Build the evaluation key from a resource disjoint from the injected one.** Here we injected from NER output and scored against expert annotations; once separated, an independence rate can be reported.
2. **Where separation is impossible, report KB coverage** and the distortion projected from it, so readers can discount the reported figure accordingly.

3. **Do not report an entity metric as a single number.** Separate the segment whose gold overlaps the system's resources from the segment that does not; this alone reproduces the difference-in-differences of §5.2.
4. **When a weak model shows a large gain, check for a ceiling effect first.** If the gain approximates the complement of prior performance, it is likely an artefact of the measurement rather than an effect of the method.
5. **Do not compare absolute performance across models with a KB-derived gold.** As §5.5 shows, such a gold penalises the more competent model more heavily.

### 9.4 Relation to adjacent work in the field

Applying large language models to cultural-heritage material and dissecting the results qualitatively is by now an established line: a generative model for classical Chinese (Liu et al., 2024) and domain-specific entity recognition for archaeology (Brandsen et al., 2022), both published in this journal. Our contribution to that line is **methodological**: rather than adding a performance figure, we ask under what conditions such figures can be trusted.

---

## 10. Future Work

**(1) Extending gold coverage.** Our distortion estimate comes from a setting with 70% KB coverage. The projection that lower-coverage domains are more exposed is a linear extrapolation and should be measured directly in a low-coverage archive.

**(2) Broader entity types.** We treat person names only. Official titles, place names and book titles fail differently and have different KB coverage, so how the loop scales by entity type is the next question.

**(3) A longer model axis.** Four models give $r = -0.938$, but $n=4$ does not support a significance test. Ten or more would settle whether the relation is linear or saturating. Measuring instruction-following ability independently would test whether it is what sets the ceiling.

**(4) Contrast with fine-tuned systems.** A fine-tuned model receives no injection, so no loop arises. Scoring one on the same evaluation set answers a separate question: **does fine-tuning solve the entity problem?** If surface fluency rises while entity accuracy does not, the case for KB injection is strengthened rather than weakened.

**(5) Transfer to other low-resource domains.** The design — separating injected from scoring resources, fixing the denominator, differencing by segment — is not specific to the *Seungjeongwon Ilgi*. It applies wherever an expert gold is unavailable and a dictionary or KB is substituted: historical archives, clinical records, legal corpora.

**(6) Standardising the protocol.** Reporting an entity metric as a single number is itself the problem. A protocol separating the segment that overlaps the system's resources from the segment that does not should be the default, and applying it retrospectively to existing benchmarks would show how many reported figures require reinterpretation.

**(7) Hardening the scoring protocol, with adversarial validation.** Our measurements use the same presence-based scoring as prior work (`gold ∈ output`), because changing the metric would confound whether the observed inflation comes from the protocol or from the loop. That scoring is, however, vulnerable to the over-generation inflation identified by KoBE (Gekhman et al., 2020) and to the "append the term at the end" evasion demonstrated by Alam et al. (2021). The next step is to define a protocol combining an over-generation penalty with a positional check, and to validate it against an **adversarial system that simply concatenates the injected names to the end of the translation**. All measurements in §5 should then be repeated under the hardened protocol to show that the conclusions do not depend on the scoring rule.

---

## 11. References

### Korean-language sources

### International sources